\documentclass{article}

\usepackage{PRIMEarxiv}

\usepackage[utf8]{inputenc} 
\usepackage[T1]{fontenc}    
\usepackage{hyperref}       
\usepackage{url}            
\usepackage{booktabs}       
\usepackage{amsfonts}       
\usepackage{nicefrac}       
\usepackage{microtype}      
\usepackage{lipsum}
\usepackage{fancyhdr}       
\usepackage{graphicx}       
\graphicspath{{media/}}     
\usepackage{amsmath}
  \usepackage[utf8]{inputenc}
\usepackage[T1]{fontenc}
\usepackage{xcolor}
\usepackage{booktabs}
\usepackage{multirow}
\usepackage{makecell}
\usepackage{pifont}
\usepackage{authblk}

\definecolor{ForestGreen}{RGB}{34,139,34}
\definecolor{BrickRed}{RGB}{203,65,84}

\usepackage{titlesec}

\titlespacing*{\section}
  {0pt}        
  {3pt}        
  {2pt}        

\titlespacing*{\subsection}
  {0pt}
  {3pt}
  {2pt}
\usepackage{caption}
\newcommand{\cmark}{\ding{51}} 
\newcommand{\xmark}{\ding{55}} 

\newcommand{\up}[1]{\textcolor{green!50!black}{\scriptsize$(\uparrow\!$#1$)$}}
\newcommand{\dn}[1]{\textcolor{red!60!black}{\scriptsize$(\downarrow\!$#1$)$}}
\newcommand{\nc}{\textcolor{gray!70!black}{\scriptsize$(=)$}}
\title{Parameter-efficient Prompt Tuning of Vision Foundation Model With Adaptive Focal Loss for Interpretable MCI Screening}

\author[1]{Javad Khoramdel\thanks{j.khorramdel96@gmail.com, mohammadhosini@email.kntu.ac.ir, a.nikoofard@kntu.ac.ir}}
\author[1]{Farhad Hoseyni}
\author[1]{Amirhossein Nikoofard}

\affil[1]{APAC Research Group, Faculty of Electrical Engineering, K. N. Toosi University of Technology, Tehran, Iran}

\begin{document}
\maketitle

\begin{abstract}
Mild Cognitive Impairment is a critical early stage of cognitive decline that frequently precedes Alzheimer's disease, yet its automated detection from neuropsychological drawing tests remains fundamentally constrained by data scarcity, class imbalance, and diagnostic ambiguity near clinical boundaries. Existing methodologies attempt to bypass these constraints using computationally expensive, fully fine-tuned hybrid architectures that relegate spatial explainability to a post-hoc approximation rather than an intrinsic model property. We propose a parameter-efficient framework utilizing frozen DINOv2-Small model adapted via three modality-specific learnable prompt tokens while Operating with 1.19 million trainable parameters, each token serves as a query in a shared cross-attention layer over the source image patch tokens. Crucially, spatial explainability is achieved directly through these attention maps; as a structural consequence of the architecture. Then task-conditioned embeddings fused via an attention module to quantify modality-level importance per subject. To handle boundary ambiguity, a MoCA-adapted focal loss introduced that integrates continuous cognitive scores into the training target, loss modulation, and adaptive sample weighting, strictly generalizing standard soft-label approaches. Under stratified five-fold cross-validation, the proposed architecture yields an MCI-class F1 of $0.641 \pm 0.026$ and an AUC of $0.795 \pm 0.024$, outperforming the computationally heavier ResViT baseline by 0.110 in MCI-class F1.
\end{abstract}


\section{Introduction}

 Mild Cognitive Impairment (MCI) occupies a clinically critical window between normal
ageing and dementia: subjects scoring below 25 on the Montreal Cognitive
Assessment (MoCA)~\cite{nasreddine2005montreal} are classified as MCI, yet
scores cluster densely around this cutoff, making the boundary inherently
uncertain and the diagnostic task sensitive to both labelling strategy and
evaluation design.  Ruengchaijatuporn et al.~\cite{ruengchaijatuporn2022explainable}
address this by jointly processing clock drawing test (CDT), cube-copying, and trail-making images
through three fully fine-tuned VGG16~\cite{simonyan2014very} backbones and
introducing MoCA-derived soft labels to relax the hard threshold, demonstrating
that multi-modal inputs and score-informed labelling substantially improve
detection.  Interpretability is recovered post-hoc via attention
rollout~\cite{abnar2020quantifying}, and evaluation relies on random splits that
do not guarantee balanced class representation on a 2.4:1 skewed dataset.
Sirshar et al.~\cite{sirshar2026mci} subsequently propose ResViT, a parallel
hybrid of a ResNet50~\cite{he2016deep} and ViT-B/16~\cite{dosovitskiy2020image},
arguing that combining local convolutional and global transformer features
is necessary for this task.  Yet their 32-million-parameter model is evaluated
on a single held-out partition without cross-validation, and performance is
reported only in aggregate, obscuring MCI-class sensitivity under class imbalance.
We argue that such architectural complexity is unnecessary. An architecture including a self-supervised
vision foundation model already encodes rich local and global representations,
and the domain gap can be closed far more efficiently through lightweight,
task-specific adaptation rather than by stacking two large pretrained networks.
Furthermore, MoCA scores carry richer supervisory information than a single
soft-label transformation can exploit, and robust evaluation on a skewed dataset
requires stratified cross-validation with MCI-class sensitivity as the primary
optimisation target.

\begin{enumerate}
    \item \textbf{Parameter-efficient, interpretable MCI classification from
    neuropsychological drawing triplets.} Modality-specific prompt tokens adapt
    a frozen DINOv2-Small~\cite{darcet2023vision} backbone with under 6\% of
    parameters trainable, yielding spatial attention maps and modality
    importance weights as direct byproducts of inference.

    \item \textbf{MoCA-aware focal loss.} A unified loss incorporates the
    continuous MoCA score into a soft target, a MoCA-probability modulator, and
    MoCA-bin adaptive weighting, strictly generalising~\cite{ruengchaijatuporn2022explainable}.

    \item \textbf{Targeted augmentation strategies.} Image inversion,
    class-balanced sampling, type-preserving Mixup, and MoCA-neighbour drawing
    swap address the photometric domain gap and increase triplet diversity.
\end{enumerate}





\begin{figure*}[t]
    \centering
    \includegraphics[width=\linewidth, trim=0 70pt 0 0, clip=true]{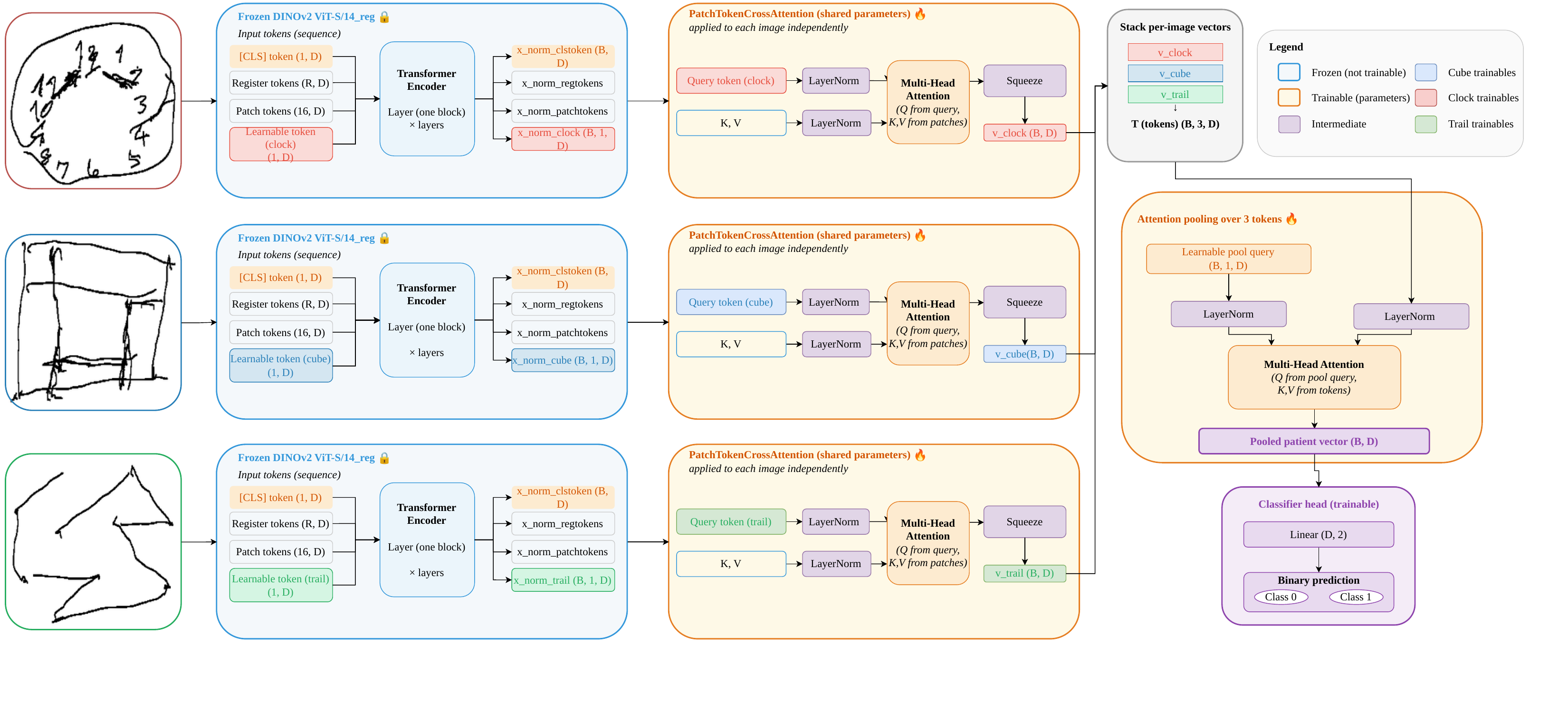}
    \caption{Overview of the proposed framework. Each drawing (clock, cube, trail) is processed
by a shared frozen DINOv2-Small backbone with a modality-specific learnable token,
which is then refined via a shared cross-attention over the backbone's
patch tokens. The 3 resulting embeddings are aggregated by a learnable query
attention pooling module into a single vector, which is passed to a linear
classifier. Frozen components are shown in blue; trainable components in orange.}
    \label{fig:model_diagram}
\end{figure*}

\section{Related Works}
\label{subsubsec:loss_prior}

Recent approaches to automated cognitive screening primarily apply deep convolutional networks (CNNs) and Vision Transformers (ViTs) to the CDT. Chen et al. \cite{chen2020automatic} combined manifold learning with DenseNet-121 for binary dementia screening, demonstrating high classification accuracy. However, their evaluation relies on a skewed dataset with a high prevalence of pathological cases, complicating cross-study generalization. Similarly, Raksasat et al. \cite{raksasat2023attentive} introduced API-Net, a contrastive learning architecture, while Liu et al. \cite{ning8improved} proposed a self-supervised contrastive framework utilizing MoCo. Both studies struggle with class imbalance. Raksasat et al. collapsed moderate and severe impairment stages into a single monolithic class, destroying fine-grained staging capability. Liu et al.'s framework suffers from catastrophic misclassifications in minority categories, resulting in severely degraded macro-averaged performance of 0.554. Hu et al. \cite{hu2026novel} extended this paradigm by applying an ordinal loss function to a ViT backbone for severity classification on a large cohort. While demonstrating strong binary discrimination, their reliance on the computationally expensive full fine-tuning of massive pre-trained networks lacks parameter efficiency.

Generative architectures have also been leveraged to extract disentangled features for CDT assessment. Bandyopadhyay et al. \cite{bandyopadhyay2023explainable} utilized a Relevance Factor Variational Autoencoder (RF-VAE) to differentiate dementia from cognitively normal patients with robust predictive performance. Addressing demographic biases inherent in such models, Zhang et al. \cite{zhang2024developing} subsequently applied static sample reweighting to mitigate elevated Type I error rates observed in patients with limited formal education. Both approaches, however, collapse the continuous spectrum of cognitive decline into rigid binary classifications.

Transitioning beyond unimodal assessment, Yang et al. \cite{yang2025multimodal} proposed a multimodal graph neural network (GNN) combining cube-copying images, demographics, and cognitive scores to achieve high diagnostic sensitivity. However, their reliance on explicit line-simplification algorithms to convert images into geometric graphs is problematic for unconstrained visuospatial tasks. Furthermore, their evaluation is constrained to a limited cohort, producing high-variance performance estimates.

Regarding loss formulations and label smoothing, the soft-label approach of Ruengchaijatuporn et al. \cite{ruengchaijatuporn2022explainable} integrates continuous cognitive scores by converting the MoCA cutoff into a soft binary cross-entropy target. The proposed framework strictly generalizes this concept. Rather than relying solely on soft targets, we introduce a MoCA-aware focal loss that couples a focal modulator with a score-probability misalignment penalty and MoCA-bin-level adaptive weighting. Furthermore, to maintain a coherent training signal under data augmentation, we apply type-preserving Mixup that linearly interpolates both the hard labels and the continuous cognitive scores, ensuring the adaptive loss components remain stable for mixed samples.

Despite these advances, existing methodologies share fundamental structural limitations. The majority remain strictly unimodal \cite{ning8improved, hu2026novel, raksasat2023attentive, chen2020automatic, bandyopadhyay2023explainable, zhang2024developing}, failing to exploit complementary diagnostic signals across diverse neuropsychological tasks. Second, spatial interpretability is overwhelmingly treated as a post-hoc approximation (e.g., CAM or SHAP) \cite{yang2025multimodal, hu2026novel, raksasat2023attentive} rather than an intrinsic architectural property. Moreover, Most of the previous works focus on fully finetuning large models, which is not computationally efficient. Finally, prior architectures typically force continuous cognitive decline into rigid classifications. The proposed framework directly addresses these constraints through parameter-efficient prompt tuning of a frozen DINOv2 foundation model, enabling multimodal fusion, yielding intrinsic spatial attention maps, and preserving continuous diagnostic uncertainty.

\section{Dataset}
\label{sec:dataset}

All experiments use the publicly available multi-drawing MCI dataset introduced
by Ruengchaijatuporn et al.~\cite{ruengchaijatuporn2022explainable}, collected
under institutional review board approval at King Chulalongkorn Memorial Hospital,
Bangkok, Thailand.  The dataset comprises 918 subjects drawn from a healthy
elderly cohort with a median age of 67 years (range 55--89), of whom 77\% were
female.  Each subject completed a digital version of the Montreal Cognitive
Assessment (MoCA) administered on a tablet with a digital stylus, and
simultaneously performed three neuropsychological drawing tasks: the Clock Drawing
Test (CDT), the Cube Copying Test (CCT), and the Trail Making Test (TMT).  Each
task was recorded as a rasterised image, yielding a triplet of drawings per subject
and 2,754 images in total.

\begin{figure}[t]
    \centering
    \begin{minipage}{0.48\linewidth}
        \centering
        \includegraphics[width=\linewidth]{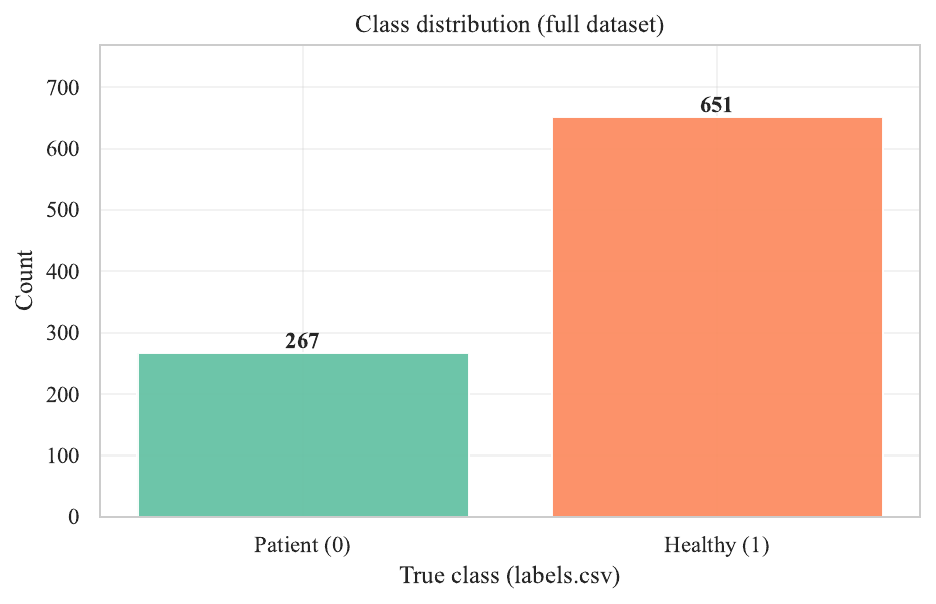}
        \caption{Class distribution of the full dataset. Of 918 subjects,
        651 are healthy controls and 267 are MCI patients. 
        }
        \label{fig:class_dist}
    \end{minipage}\hfill
    \begin{minipage}{0.48\linewidth}
        \centering
        \includegraphics[width=\linewidth]{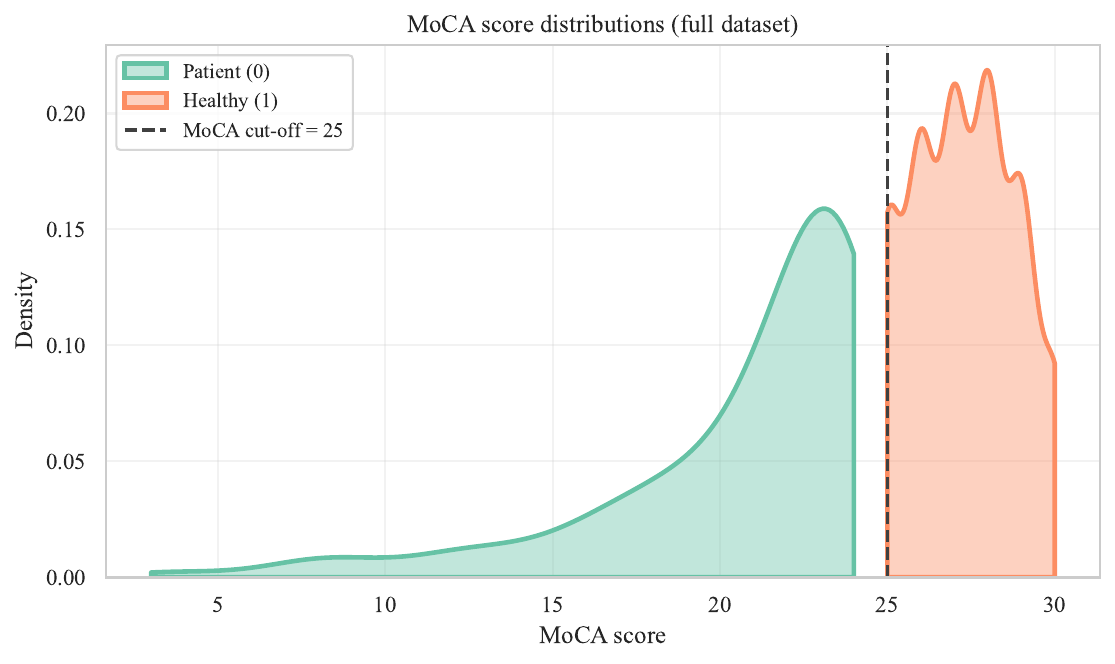}
        \caption{MoCA-score distribution across the dataset, with a mode of 27. The dashed
        line marks the clinical cutoff of 25.}
        \label{fig:moca_dist}
    \end{minipage}
\end{figure}

Subjects are labelled according to the standard clinical criterion: those with a
MoCA score of 25 or above are classified as healthy controls (HC) and those with a
score below 25 as MCI patients.  This results in 651 HC subjects and 267 MCI
patients, a 2.4:1 class ratio, as illustrated in Figure~\ref{fig:class_dist}.
The MoCA score distribution across the full dataset is shown in
Figure~\ref{fig:moca_dist}.  The distribution is unimodal and right-skewed, with
the mode falling near a score of 27 and a long left tail extending toward the
lower end of the scale.  A notable concentration of subjects falls immediately
above and below the cutoff of 25, forming a region of inherent diagnostic
ambiguity that motivates the MoCA-aware loss formulation described in
Section~\ref{subsec:loss}. Five-fold cross-validation is employed throughout, with folds stratified by binary
label to preserve the class ratio in each split.  


\section{Methodology}
\label{sec:methodology}

We propose a parameter-efficient, interpretable framework for detecting Mild Cognitive
Impairment (MCI) from three neuropsychological drawing tests: the Clock Drawing Test
(CDT), the Cube Copying Test (CCT), and the Trail Making Test (TMT).  The framework
comprises three principal contributions.  First, a \emph{task-specific prompt-tuned}
DINOv2 backbone extracts a dedicated representation for each drawing test while keeping
all pre-trained weights frozen, achieving adaptation at a fraction of the parameters
required by full fine-tuning.  Second, a \emph{learnable query attention} fusion module
aggregates the three task-level embeddings into a single vector for binary
classification, providing an interpretable, subject-level importance score over the
three drawing modalities.  Third, a \emph{MoCA-aware focal loss} incorporates each
subject's continuous cognitive score into both the training target and the per-sample
loss modulation, enabling the model to explicitly account for diagnostic uncertainty
near the clinical boundary.  In addition, a set of targeted training and augmentation
strategies (such as: image inversion, class-balanced batch sampling, type-preserving Mixup, and
MoCA-neighbour drawing swap) are introduced to address the photometric domain gap and
the limited subject-level diversity characteristic of neuropsychological drawing
datasets.  

\subsection{preleminary}
The proposed architecture consists of a foundation model as the backbone for vision feature extraction. To enhance the performance of the architecture for downstream tasks, the backbone model should be fine-tuned. For this purpose, Parameter-efficient fine-tuning (PEFT) methodologies are used as modern techniques to adapt large-scale foundation models to downstream tasks while mitigating severe computational and storage overhead. To establish a comparative baseline, the fine-tuning of top layers (top tuning)~\cite{alfano2024top} is frequently applied, albeit requiring the optimization of a comparatively larger subset of task-specific parameters. Alternatively, Low-Rank Adaptation (LoRA)~\cite{hu2022lora} is implemented by freezing the original pre-trained network weights and exclusively optimizing injected low-rank decomposition matrices, thereby enabling practical and efficient model adaptation. Furthermore, prompt tuning strategies specifically engineered for vision-language models, such as Context Optimization (CoOp)~\cite{zhou2022learning}, are deployed by modeling prompt context words as continuous, learnable vectors, successfully aligning fixed model representations with new visual categories without updating the underlying image or text encoders.

\subsection{Parameter-Efficient Feature Extraction via Task-Specific Prompt Tuning}
\label{subsec:architecture}

The DINOv2-Small with registers~\cite{oquab2023dinov2,darcet2023vision} is adopted as the visual backbone.  DINOv2 is a Vision
Transformer~\cite{dosovitskiy2020image} trained through self-supervised
knowledge distillation on large-scale curated image collections, yielding
semantically rich, transferable patch-level representations without relying
on labelled data.  The \emph{registers} variant augments the standard
sequence with $R$ additional register tokens that absorb high-norm patch
outliers, yielding cleaner local attention maps~\cite{darcet2023vision} and
directly benefiting the spatial interpretability. All backbone parameters are
frozen throughout training.  The overall architecture is illustrated in
Figure~\ref{fig:model_diagram}.

\subsubsection{Task-Specific Prompt Tokens and Cross-Attention Aggregation}
\label{subsubsec:prompt_tokens}

The three drawing tests probe distinct cognitive functions: visuospatial
planning and executive function (CDT), three-dimensional construction (CCT),
and visual scanning together with cognitive flexibility
(TMT)~\cite{borson2003mini}.  To capture these distinctions, we introduce a
dedicated learnable prompt token: $t_{k} \in R^{d}, k \in \{\text{clock}, \text{cube}, \text{tail}\}$;
initialized randomly and optimized end-to-end.  For task $k$, the frozen
patch-embedding layer partitions image $\mathbf{x}_k$ into $N$
non-overlapping $14{\times}14$-pixel patches and projects each to
$\mathbb{R}^d$, yielding patch tokens $\{\mathbf{p}_1^k,\ldots,\mathbf{p}_N^k\}$.
These propagate through the $L$ frozen transformer encoder blocks of DINOv2
alongside the built-in class token and $R$ register tokens, producing
final-layer patch representations $\{\tilde{\mathbf{p}}_1^k,\ldots,
\tilde{\mathbf{p}}_N^k\}$.

\paragraph{Task-specific cross-attention.}
Rather than reading the final-layer CLS token directly, each task-specific
token $\mathbf{t}_k$ is used as the \emph{query} in a cross-attention
operation whose keys and values are the final-layer patch tokens of the
corresponding image:
\begin{equation}
    \mathbf{e}_k
    \;=\;
    \mathrm{CrossAttn}\!\Bigl(
        \underbrace{\mathbf{t}_k \mathbf{W}_Q^{\mathrm{x}}}_{\text{query}},\;
        \underbrace{\tilde{\mathbf{P}}_k \mathbf{W}_K^{\mathrm{x}}}_{\text{keys}},\;
        \underbrace{\tilde{\mathbf{P}}_k \mathbf{W}_V^{\mathrm{x}}}_{\text{values}}
    \Bigr)
    \;\in\; \mathbb{R}^{d},
    \label{eq:cross_attn}
\end{equation}
where $\tilde{\mathbf{P}}_k \in \mathbb{R}^{N \times d}$ is the stacked
matrix of final-layer patch tokens and
$\mathbf{W}_Q^{\mathrm{x}}, \mathbf{W}_K^{\mathrm{x}},
\mathbf{W}_V^{\mathrm{x}}$ are learned projection matrices
\emph{shared across all three tasks}.  This design programmes each token
$\mathbf{t}_k$ to learn which spatial patches of its assigned drawing are
most informative for the corresponding cognitive function, producing a
task-conditioned summary embedding $\mathbf{e}_k$.

\paragraph{Spatial interpretability.}
The cross-attention weights from $\mathbf{t}_k$ to each patch constitute a
natural spatial saliency map. Averaging over heads and reshaping to
$\sqrt{N}\times\sqrt{N}$:
\begin{equation}
    \mathbf{A}_k
    \;=\;
    \mathrm{Reshape}\!\left(
        \frac{1}{H}\sum_{h=1}^{H} \alpha_{h,k,n}^{\mathrm{x}}
    \right)_{n=1}^{N}
    \;\in\; \mathbb{R}^{\sqrt{N}\times\sqrt{N}},
    \label{eq:attn_map}
\end{equation}
where $\alpha_{h,k,n}^{\mathrm{x}}$ is the attention weight from token
$\mathbf{t}_k$ to patch $n$ in head $h$.  These maps are a direct product
of the forward pass and require no additional backward computation, unlike
gradient-based attribution
methods~\cite{selvaraju2017grad,rafati2025benchmarking}. 

\subsection{Multi-Task Fusion via Learnable Query Attention}
\label{subsec:fusion}

The three cross-attention outputs are stacked as
$\mathbf{E} = [\mathbf{e}_1;\mathbf{e}_2;\mathbf{e}_3] \in \mathbb{R}^{3
\times d}$ and aggregated by a learnable query attention
mechanism~\cite{lee2019set}.  A single learnable query vector
$\mathbf{q} \in \mathbb{R}^{d_q}$ attends over $\mathbf{E}$:
\begin{equation}
    \mathbf{f}
    \;=\;
    \underbrace{
        \mathrm{softmax}\!\left(
            \frac{(\mathbf{q}\mathbf{W}_Q)\,(\mathbf{E}\mathbf{W}_K)^\top}
                 {\sqrt{d_k}}
        \right)
    }_{\boldsymbol{\alpha}\,\in\,\Delta^2}
    (\mathbf{E}\mathbf{W}_V)
    \;\in\; \mathbb{R}^{d_v},
    \label{eq:attention_pool}
\end{equation}
where $\boldsymbol{\alpha}=(\alpha_1,\alpha_2,\alpha_3)$ lies on the
probability simplex $\Delta^2$.  This vector forms a second, complementary
interpretability signal: while $\mathbf{A}_k$ reveals \emph{where} the
model attends within each drawing, $\boldsymbol{\alpha}$ reveals
\emph{which} drawing test most influenced the classification decision for a
given subject, which may assist a clinician in identifying the cognitive
domain most implicated in a subject's risk profile.  The fused vector
$\mathbf{f}$ is passed through a linear classifier.

\subsection{MoCA-Adapted Focal Loss}
\label{subsec:loss}
Standard BCE assigns identical weight to every sample and imposes a hard label at a
fixed MoCA cutoff, which ignores diagnostic uncertainty near the clinical boundary and
is sensitive to class imbalance. Both of them are addressed with a unified \emph{MoCA-adapted focal
loss} that controls (i) the effective training target, (ii) the loss modulation
strategy, and (iii) per-sample weighting by MoCA-bin frequency.




\subsubsection{Notation and Soft Anchor}
Let $p_i = \mathrm{softmax}(\mathbf{z}_i)_{\text{HC}} \in [\varepsilon, 1-\varepsilon]$
be the predicted HC probability for subject $i$, with $m_i \in [0,30]$ their MoCA
score, $y_i \in \{0,1\}$ their hard label ($y_i=1$ iff $m_i \geq 25$), and $C=25.5$
the default clinical cutoff. The \emph{MoCA soft anchor} is defined as:
\begin{equation}
    s_i \;=\; \sigma\!\bigl(m_i - C\bigr) \;\in\; (0,1),
    \label{eq:soft_anchor}
\end{equation}
where $\sigma$ is the sigmoid function. The
effective training target is $y_{\mathrm{eff},i} = s_i$ if soft targets are enabled,
and $y_{\mathrm{eff},i} = y_i$ otherwise.

\subsubsection{Loss Modulators}
Two modulator families are proposed that reweight each log-likelihood term independently.

\paragraph{Classic focal modulator.} Following~\cite{lin2017focal}, the modulators
down-weight confident, easy samples via $m_{1,i} = (1-p_i)^{\gamma}$ and
$m_{2,i} = p_i^{\gamma}$, with focusing parameter $\gamma \geq 0$. Setting $\gamma=0$
recovers standard BCE.

\paragraph{MoCA–probability modulator.} A novel alternative is introduced that focuses
training on samples where the model's prediction is \emph{inconsistent with the
MoCA-derived expectation}:
\begin{equation}
    m_{1,i} \;=\; m_{2,i} \;=\; \bigl(|s_i - p_i| + \varepsilon\bigr)^{b},
    \label{eq:moca_mod}
\end{equation}
with exponent $b \geq 0$. When $p_i \approx s_i$, the modulator is small and the
sample contributes little to the loss; when the two diverge, the modulator amplifies
that sample's gradient. Unlike the classic focal modulator, this is explicitly tied to
the subject's cognitive score rather than prediction confidence alone.

\subsubsection{Adaptive Per-Sample Weight}
To correct for unequal MoCA-bin frequencies, each sample receives an
\emph{effective-number weight}:
\begin{equation}
    w_i \;=\; \frac{1 - \gamma_w}{1 - \gamma_w^{N_b}},
    \label{eq:eff_weight}
\end{equation}
where $N_b$ is the number of training samples in MoCA bin $b$ (scores rounded to the
nearest integer) and $\gamma_w \in (0,1)$ (default $0.99$). Over-represented bins are
down-weighted and rare bins up-weighted, simultaneously correcting for class imbalance
and MoCA-score skewness. When adaptive weighting is disabled, $w_i = 1$.

\subsubsection{Unified Loss}
The per-sample and batch losses are
\begin{equation}
    \ell_i \;=\; -\,w_i\Bigl(
        m_{1,i}\,y_{\mathrm{eff},i}\log p_i
        + m_{2,i}(1-y_{\mathrm{eff},i})\log(1-p_i)
    \Bigr), \qquad
    \mathcal{L} \;=\; \frac{1}{B}\sum_{i=1}^{B}\ell_i.
    \label{eq:per_sample_loss}
\end{equation}
Table~\ref{tab:loss_configs} summarises the four configurations evaluated in our
experiments.
\begin{table}[h]
\centering
\caption{Loss configurations compared in experiments. BCE = binary cross-entropy;
$\gamma$, $b$ are the focal and MoCA–probability exponents respectively. All
configurations use the same classifier and fusion module.}
\label{tab:loss_configs}
\setlength{\tabcolsep}{6pt}
\begin{tabular}{lcccc}
\toprule
\textbf{Configuration} & Soft Target & MoCA Mod. & Modulator & Effective target \\
\midrule
Hard BCE
    & \xmark & \xmark & $(1-p)^0,\,p^0=1$  & $y_i\in\{0,1\}$ \\
Soft-label BCE~\cite{ruengchaijatuporn2022explainable}
    & \cmark & \xmark & $1$                 & $s_i$ \\
Hard + MoCA mod.\ (ours)
    & \xmark & \cmark & $|s_i-p_i|^b$       & $y_i\in\{0,1\}$ \\
Soft + MoCA mod.\ (ours)
    & \cmark & \cmark & $|s_i-p_i|^b$       & $s_i$ \\
\bottomrule
\end{tabular}
\end{table}

\subsection{Training Augmentation and Sampling Strategies}
\label{subsec:tricks}
Four complementary techniques address the domain gap between neuropsychological
drawings and natural-image pre-training data, class imbalance, and limited dataset
diversity. Each is independently configurable and ablated in
Section~\ref{sec:experiments}.

\paragraph*{Image Inversion} Since neuropsychological drawings consist of dark strokes on a white background (the inverse of the natural-image statistics DINOv2 was pre-trained on) we optionally
invert pixel intensities ($\tilde{x} = 255 - x$) before any transforms. This maps
background pixels to zero and makes stroke pixels the dominant activations, reducing
the photometric domain gap.

\paragraph*{Class-Balanced Batch Sampling}
To counteract class imbalance, we optionally draw $B/2$ samples from each class per
batch, complementing the MoCA-bin adaptive weights of Equation~\eqref{eq:eff_weight}.

\paragraph*{Type-Preserving Mixup}
To prevent clinically meaningless cross-modal blending, each drawing type is mixed
only with another sample of the same type. A single coefficient
$\lambda \sim \mathrm{Beta}(\alpha_{\mathrm{mix}}, \alpha_{\mathrm{mix}})$ is shared
across all three types, so the same subject pair is mixed in every modality.

\paragraph*{MoCA-Neighbour Drawing Swap}
\label{subsubsec:swap}
To increase structural triplet diversity beyond pixel-level interpolation, we
introduce \emph{drawing swap}: with probability $p_{\mathrm{swap}}$, one or two
drawings of a subject are replaced by the corresponding drawings of a MoCA-score
nearest neighbour. At least one drawing always originates from the base subject,
and the synthetic triplet inherits its label and MoCA score. The rationale is that
MoCA-similar subjects share comparable cognitive deficit patterns, so the resulting
triplet remains diagnostically coherent. Drawing swap is applied only to MCI patients
during training.

\section{Experiments}
\label{sec:experiments}
 
\subsection{Experimental Setup}
\label{subsec:setup}
 
All experiments are conducted under a five-fold stratified cross-validation protocol.
The folds are generated from a fixed random seed (42) and kept constant across every
configuration to ensure strictly comparable results. Within each fold, the checkpoint
that achieves the best MCI-class F1 score on the held-out validation split is retained
for evaluation. We report the mean and standard deviation of each metric across the five
folds. All configurations share the same optimization schedule: AdamW~\cite{loshchilov2017decoupled}
with a learning rate of $10^{-4}$, weight decay $0.05$, and a three-epoch linear
warm-up followed by cosine annealing, for a total of 40 training epochs. The batch size
is 4. A common data augmentation pipeline is applied in every configuration, comprising
random translations (spatial fraction 0.03), additive Gaussian noise ($\sigma = 0.02$),
random Gaussian blur (kernel $\in \{3,\,5\}$, application probability 0.25), and colour
jitter (brightness, contrast, and saturation $\pm 0.15$; hue $\pm 0.03$).
 
\paragraph{Metrics.}
MCI screening is a safety-critical task in which missed diagnoses carry greater clinical
cost than false alarms. We therefore designate the MCI-class F1 score
($F1_{\mathrm{MCI}}$) as the primary metric for checkpoint selection and configuration
ranking. To characterise each model's precision-recall operating point more completely,
we additionally report the area under the ROC curve (AUC-ROC), overall accuracy,
macro-averaged F1 (Macro-F1), and MCI-class recall and precision.
 
\subsection{Individual Contribution of Training Strategies}
\label{subsec:individual}
 
Table~\ref{tab:comparison} evaluates each proposed training strategy in isolation: a
single technique is added to the base model while every other setting remains at its
default.  The base model uses hard binary cross-entropy loss, the task-specific
prompt-tuned DINOv2 architecture described in Section~\ref{subsec:architecture}, and no
components beyond the shared augmentation pipeline. We also include a cross-architecture comparison against the reference method
of~\cite{sirshar2026mci}. Because the authors do not release source code, we
re-implement their architecture exactly as described in the paper and train it under
the same conditions as our base model.  We additionally report a row corresponding to
the MoCA Soft Target strategy proposed in~\cite{ruengchaijatuporn2022explainable}, which our framework
strictly generalises as shown in Section~\ref{subsubsec:loss_prior}; this entry
measures the performance of that strategy alone on our architecture.
 
\paragraph{Observations.}
All single-technique variants improve $F1_{\mathrm{MCI}}$ over the base model.
Type-Preserving Mixup achieves the largest gain in $F1_{\mathrm{MCI}}$ ($+0.022$) and
Macro-F1 ($+0.020$) among all individual techniques, and ranks second in AUC-ROC
($+0.018$). The MoCA Soft Target attains the highest AUC-ROC across all single-strategy
configurations ($+0.019$), indicating that MoCA-guided label smoothing improves the
model's ranking of subjects across classification thresholds, consistent with its role
in relaxing overconfident predictions near the diagnostic boundary.
 
Image Inversion yields the highest accuracy ($0.788$) and the largest improvement in
MCI-class precision ($+0.042$), suggesting that mapping ink strokes to bright foreground
on a black background better aligns the drawing images with the photometric statistics of
the natural-image data on which DINOv2 was pre-trained, and directs patch-level
representations toward diagnostically relevant stroke regions rather than the
uninformative white canvas. Balanced Batch Sampling improves accuracy and precision by
correcting the class-level imbalance in the training stream. Adaptive MoCA-Bin Weighting
achieves the highest MCI-class recall of any single technique ($+0.071$), at the cost of
precision ($-0.037$); this trade-off is expected, as the effective-number reweighting
explicitly elevates the influence of under-represented MoCA bins, which are
disproportionately populated by MCI subjects. Drawing Swap provides consistent
improvements in $F1_{\mathrm{MCI}}$ and precision, albeit slightly lower in magnitude
than Mixup, reflecting its complementary role in expanding the combinatorial diversity of
multi-modal training triplets. The MoCA-Probability Modulator offers modest but
consistent gains across most metrics.

When all strategies are combined into the full system, the model achieves the
highest $F1_{\mathrm{MCI}}$ across the entire table ($0.641$), demonstrating that the
individual improvements are largely complementary rather than redundant, with the
primary gain concentrated in MCI-class recall and F1. The reference architecture of~\cite{sirshar2026mci} underperforms the base model across
every metric. Its $F1_{\mathrm{MCI}}$ falls $0.086$ points below the base model and its
MCI-class recall is $0.135$ points lower, confirming that the representational power of
the frozen DINOv2 backbone combined with task-specific prompt tuning provides a
substantially stronger foundation for this task than the convolutional-transformer hybrid
proposed in the reference work.

\subsection{Qualitative Analysis of Attention Maps and Modality Weights}
\label{subsec:qualitative}

\begin{figure}[t]
    \centering
    \includegraphics[width=0.9\linewidth]{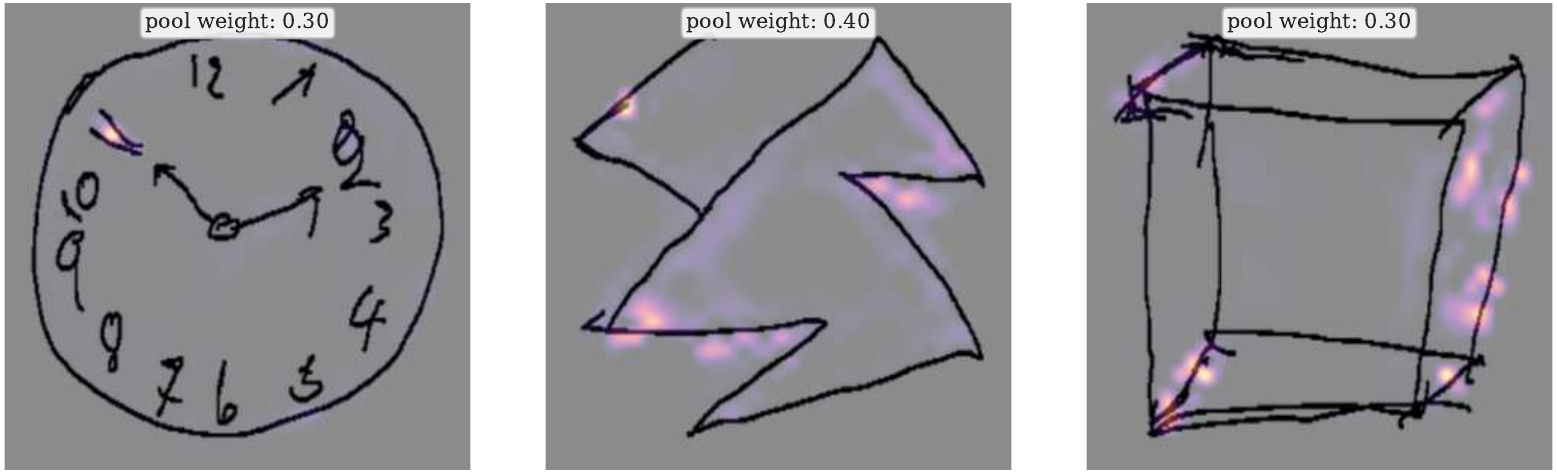}\\[6pt]
    \includegraphics[width=0.9\linewidth]{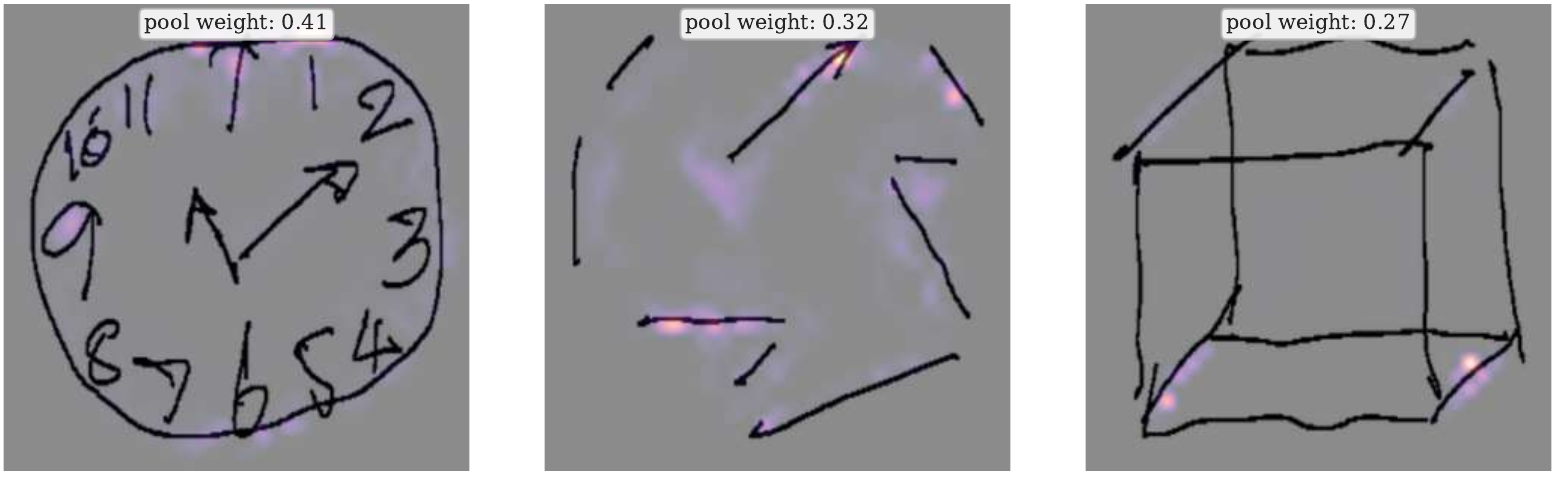}
    \caption{Cross-attention maps and learnable pool weights for two
    representative subjects.  Each column shows one drawing modality (clock,
    trail, cube from left to right); warm colours indicate high attention.
    Pool weights (shown above each panel) are the components of
    $\boldsymbol{\alpha}$ and sum to one across the three modalities.
    \textbf{Top row:} MCI patient correctly classified as MCI.
    \textbf{Bottom row:} healthy subject incorrectly classified as MCI
    (false positive).}
    \label{fig:attn_qualitative}
\end{figure}

Figure~\ref{fig:attn_qualitative} presents two representative subjects alongside
the cross-attention maps $\{\mathbf{A}_k\}$ and the learnable query attention
weights $\boldsymbol{\alpha}$ produced by the model at inference.  Together,
these two interpretability signals expose not only the spatial regions that drove
each task-specific embedding, but also the relative contribution of each drawing
modality to the final binary prediction.

\paragraph{True positive (MCI patient, top row).}
The model assigns the highest pool weight to the trail drawing
($\alpha_{\mathrm{trail}} = 0.40$), with clock and cube receiving equal weights
of 0.30.  Inspecting the trail attention map reveals that the model concentrates
on the junction points where strokes are incorrectly connected, a salient marker
of executive dysfunction in the TMT.  This is consistent with clinical scoring
practice, in which erroneous path connections are the primary error category
in MCI patients on this task.  On the clock drawing, attention localizes around
the digit positions for 2 and 11, where stroke density is elevated relative to
a correctly drawn clock, suggesting the model identifies these as anomalous
regions.  On the cube, attention highlights the corners and edge intersections
where superfluous strokes and misaligned lines accumulate, another common
indicator of visuospatial difficulty.  The modality weight pattern is therefore
clinically coherent: the trail, which shows the clearest structural error, is
assigned the most diagnostic weight.

\paragraph{False positive (healthy subject, bottom row).}
The model assigns the highest weight to the clock drawing
($\alpha_{\mathrm{clock}} = 0.41$), followed by the trail ($0.32$) and the
cube ($0.27$).  The clock attention map concentrates near the top of the face
where the digit 12 is absent or atypically rendered, and around the 9 position.
The absence of a clearly formed 12 is a well-known indicator of MCI in the CDT,
and its misinterpretation here likely drives the elevated clock weight and
contributes to the erroneous prediction.  On the trail drawing, the attention
map highlights two parallel stroke segments that are not connected to one another,
a configuration that superficially resembles an incorrect path connection.
However, in a healthy subject these isolated strokes may simply reflect a
different drawing style rather than a genuine executive failure.  This false
positive therefore illustrates a known limitation of purely image-based MCI
screening: subtle stylistic variation in an otherwise intact drawing can
activate the same spatial attention patterns as genuine impairment, particularly
near the diagnostic boundary.
 
\begin{table}[t]
  \centering
  \caption{
    Per-technique comparison against the base model under five-fold cross-validation
    (mean $\pm$ std). Coloured arrows in parentheses denote absolute change relative to
    the base model: {\color{green!50!black}$\uparrow$}~improvement,
    {\color{red!60!black}$\downarrow$}~degradation, {\color{gray!70!black}$(=)$}~no
    change. \textbf{Bold} denotes the best result and \underline{underline} the
    second-best in each column; the reference architecture is excluded from this
    ranking. $F1_{\mathrm{MCI}}$ is the primary metric.
  }
  \label{tab:comparison}
  \resizebox{\linewidth}{!}{%
  \begin{tabular}{l c c c c c c}
    \toprule
    \textbf{Configuration}
      & $F1_{\mathrm{MCI}}$ {\small (primary)}
      & AUC-ROC
      & Accuracy
      & Macro-F1
      & $\mathrm{Recall}_{\mathrm{MCI}}$
      & $\mathrm{Prec.}_{\mathrm{MCI}}$ \\
    \midrule
    Base Model
      & $0.617 \pm 0.055$
      & $0.784 \pm 0.031$
      & $0.764 \pm 0.063$
      & $0.722 \pm 0.054$
      & $0.648 \pm 0.094$
      & $0.607 \pm 0.101$ \\
    \addlinespace
    \quad + Balanced Batch Sampling
      & $0.631 \pm 0.027$ \up{0.014}
      & $0.798 \pm 0.026$ \up{0.014}
      & $\underline{0.786 \pm 0.014}$ \up{0.022}
      & $\underline{0.740 \pm 0.010}$ \up{0.018}
      & $0.633 \pm 0.098$ \dn{0.015}
      & $\underline{0.645 \pm 0.071}$ \up{0.038} \\[3pt]
    \quad + Image Inversion
      & $0.627 \pm 0.040$ \up{0.010}
      & $0.784 \pm 0.030$ \nc
      & $\mathbf{0.788 \pm 0.023}$ \up{0.024}
      & $0.739 \pm 0.024$ \up{0.017}
      & $0.617 \pm 0.099$ \dn{0.031}
      & $\mathbf{0.649 \pm 0.060}$ \up{0.042} \\[3pt]
    \quad + Type-Preserving Mixup
      & $\mathbf{0.639 \pm 0.030}$ \up{0.022}
      & $\underline{0.802 \pm 0.031}$ \up{0.018}
      & $0.783 \pm 0.013$ \up{0.019}
      & $\mathbf{0.742 \pm 0.015}$ \up{0.020}
      & $0.663 \pm 0.086$ \up{0.015}
      & $0.624 \pm 0.039$ \up{0.017} \\[3pt]
    \quad + Drawing Swap
      & $0.632 \pm 0.027$ \up{0.015}
      & $0.784 \pm 0.039$ \nc
      & $0.784 \pm 0.021$ \up{0.020}
      & $\underline{0.740 \pm 0.021}$ \up{0.018}
      & $0.637 \pm 0.045$ \dn{0.011}
      & $0.631 \pm 0.045$ \up{0.024} \\[3pt]
    \quad + MoCA Soft Target~\cite{ruengchaijatuporn2022explainable}
      & $\underline{0.638 \pm 0.044}$ \up{0.021}
      & $\mathbf{0.803 \pm 0.027}$ \up{0.019}
      & $0.777 \pm 0.034$ \up{0.013}
      & $0.738 \pm 0.033$ \up{0.016}
      & $\underline{0.678 \pm 0.081}$ \up{0.030}
      & $0.611 \pm 0.066$ \up{0.004} \\[3pt]
    \quad + Adaptive MoCA-Bin Weighting
      & $0.622 \pm 0.053$ \up{0.005}
      & $0.784 \pm 0.027$ \nc
      & $0.742 \pm 0.069$ \dn{0.022}
      & $0.711 \pm 0.056$ \dn{0.011}
      & $\mathbf{0.719 \pm 0.114}$ \up{0.071}
      & $0.570 \pm 0.123$ \dn{0.037} \\[3pt]
    \quad + MoCA-Probability Modulator
      & $0.623 \pm 0.053$ \up{0.006}
      & $0.782 \pm 0.033$ \dn{0.002}
      & $0.772 \pm 0.046$ \up{0.008}
      & $0.729 \pm 0.043$ \up{0.007}
      & $0.644 \pm 0.083$ \dn{0.004}
      & $0.614 \pm 0.082$ \up{0.007} \\
      \addlinespace
\quad Full System (all strategies combined)
  & $\mathbf{0.641 \pm 0.026}$ \up{0.024}
  & $0.795 \pm 0.024$ \up{0.011}
  & $0.765 \pm 0.042$ \up{0.001}
  & $0.732 \pm 0.031$ \up{0.010}
  & $\mathbf{0.719 \pm 0.096}$ \up{0.071}
  & $0.593 \pm 0.086$ \dn{0.014} \\
    \midrule
    \textit{Reference architecture}~\cite{sirshar2026mci}
      & $0.531 \pm 0.045$ \dn{0.086}
      & $0.741 \pm 0.037$ \dn{0.043}
      & $0.741 \pm 0.019$ \dn{0.023}
      & $0.675 \pm 0.018$ \dn{0.047}
      & $0.513 \pm 0.116$ \dn{0.135}
      & $0.568 \pm 0.043$ \dn{0.039} \\
    \bottomrule
  \end{tabular}}
\end{table}

\section{Ablation Study}
\label{sec:ablation}

We conduct three targeted ablation studies.  The first evaluates the
MoCA-adapted focal loss by removing one component at a time from the full
system.  The second compares prompt-token design choices to isolate the
architectural contribution of task-specific adaptation.  The third compares
prompt tuning against alternative parameter-efficient fine-tuning strategies
to justify the choice of backbone adaptation mechanism.

\subsection{Leave-One-Out Ablation of MoCA-adapted Loss Components}
\label{subsec:ablation_loss}

Table~\ref{tab:ablation} reports a leave-one-out analysis of the three
MoCA-adapted loss components: the MoCA-probability modulator, adaptive
MoCA-bin weighting, and MoCA soft labels.  In each row, one component is
removed while all remaining strategies are kept active.  This design ensures that the measured
degradation reflects the unique contribution of the removed component rather
than any confound introduced by simultaneously changing other settings.

A central motivation for this analysis is the relationship established in
Table~\ref{tab:loss_configs}: MoCA soft labels, as proposed
in~\cite{ruengchaijatuporn2022explainable}, constitute a strict special case
of the proposed MoCA-adapted focal loss, obtained by disabling both the
modulator and adaptive weighting.  The ablation therefore directly quantifies
how much the additional components contribute \emph{beyond} the soft-label
baseline.

\begin{table*}[t]
\centering
\caption{%
  Leave-one-out ablation of MoCA-adapted loss components within the full
  system.  Results are mean $\pm$ std across five folds.  Parenthesised
  values show the absolute change relative to the full system:
  {\textcolor{ForestGreen}{($\uparrow$\,value)}} denotes improvement and
  {\textcolor{BrickRed}{($\downarrow$\,value)}} denotes degradation.
  \textbf{Bold} marks the best result per column;
  \underline{underline} marks the second best.%
}
\label{tab:ablation}
\resizebox{\linewidth}{!}{%
\setlength{\tabcolsep}{6pt}
\renewcommand{\arraystretch}{1.30}
\begin{tabular}{l cccccc}
\toprule
\textbf{Configuration}
  & \textbf{$F1_{\mathrm{MCI}}$}
  & \textbf{AUC-ROC}
  & \textbf{Macro-F1}
  & \textbf{Accuracy}
  & \textbf{Recall$_{\mathrm{MCI}}$}
  & \textbf{Prec.$_{\mathrm{MCI}}$} \\
\midrule
Full System
  & $\mathbf{0.641} \pm 0.026$
  & $\underline{0.795} \pm 0.024$
  & $\mathbf{0.732} \pm 0.031$
  & $\underline{0.765} \pm 0.042$
  & $0.719 \pm 0.096$
  & $\underline{0.593} \pm 0.086$ \\[2pt]
\quad w/o MoCA-Probability Modulator
  & $0.624 \pm 0.050$ \dn{0.017}
  & $0.778 \pm 0.044$ \dn{0.017}
  & $0.706 \pm 0.055$ \dn{0.026}
  & $0.731 \pm 0.066$ \dn{0.034}
  & $\mathbf{0.760} \pm 0.104$ \up{0.041}
  & $0.545 \pm 0.097$ \dn{0.048} \\[2pt]
\quad w/o Adaptive MoCA-Bin Weighting
  & $\underline{0.634} \pm 0.019$ \dn{0.007}
  & $0.789 \pm 0.031$ \dn{0.006}
  & $\underline{0.722} \pm 0.023$ \dn{0.010}
  & $0.751 \pm 0.031$ \dn{0.014}
  & $\underline{0.741} \pm 0.068$ \up{0.022}
  & $0.560 \pm 0.049$ \dn{0.033} \\[2pt]
\quad w/o MoCA Soft Labels
  & $0.636 \pm 0.026$ \dn{0.005}
  & $\mathbf{0.802} \pm 0.018$ \up{0.007}
  & $0.730 \pm 0.034$ \dn{0.002}
  & $\mathbf{0.766} \pm 0.047$ \up{0.001}
  & $0.701 \pm 0.124$ \dn{0.018}
  & $\mathbf{0.600} \pm 0.083$ \up{0.007} \\
\bottomrule
\end{tabular}%
}
\end{table*}

Removing the MoCA-probability modulator produces the most severe degradation
across the primary metrics, with $F1_{\mathrm{MCI}}$ declining by 0.017
points, AUC-ROC by 0.017 points, and Macro-F1 by 0.026 points, the largest
drops observed in this analysis.  This finding confirms that directly
penalizing the misalignment between the model's predicted probability and the
MoCA-derived expectation provides a training signal that is not redundant with
either soft labels or adaptive weighting.  The simultaneous rise in MCI recall
($+0.041$) upon removing this component suggests that its primary effect is to
regularize the decision boundary, preventing the model from compensating for
difficult borderline samples through indiscriminate threshold shifting.

Removing adaptive MoCA-bin weighting produces a moderate but consistent
degradation in $F1_{\mathrm{MCI}}$ ($-0.007$), Macro-F1 ($-0.010$), and
accuracy ($-0.014$), confirming that fine-grained score-distribution
reweighting provides a useful complement to the class-level correction applied
by balanced batch sampling.

The ablation of MoCA soft labels yields a more nuanced profile: AUC-ROC
increases by 0.007 points and accuracy by 0.001 points relative to the full
system, yet $F1_{\mathrm{MCI}}$ and recall both decline.  This pattern is
consistent with the theoretical role of soft labels: relaxing the hard
boundary near the MoCA cutoff improves sensitivity at a modest cost to
overall confidence calibration as captured by AUC-ROC.

Crucially, the degradation from removing the MoCA-probability modulator
($-0.017$ in $F1_{\mathrm{MCI}}$) exceeds the degradation from removing
MoCA soft labels ($-0.005$), even though the latter is the richer component
in terms of the information it encodes.  This ordering is consistent with the
individual strategy results in Table~\ref{tab:comparison}, where the
MoCA-probability modulator and adaptive weighting each produced a larger
boost in $F1_{\mathrm{MCI}}$ than MoCA soft labels when applied in isolation.
Together, these results confirm that the proposed MoCA-adapted focal loss
contributes meaningfully beyond what the soft-label formulation of prior
work~\cite{ruengchaijatuporn2022explainable} provides.

\subsection{Prompt-Token Design Ablation}
\label{subsec:ablation_prompt}

Table~\ref{tab:prompt_ablation} compares three prompt-token configurations
under identical baseline training conditions (no MoCA-adapted loss, no
augmentation beyond the shared pipeline) to isolate the architectural
contribution of task-specific prompt tuning from any training-strategy
effects.

\begin{table*}[t]
\centering
\caption{%
  Comparison of prompt-token design choices under baseline training
  conditions.  Results are mean $\pm$ std across five folds.
  \textbf{Bold} marks the best result per column;
  \underline{underline} marks the second best.%
}
\label{tab:prompt_ablation}
\resizebox{\linewidth}{!}{%
\setlength{\tabcolsep}{6pt}
\renewcommand{\arraystretch}{1.30}
\begin{tabular}{l cccccc}
\toprule
\textbf{Configuration}
  & \textbf{$F1_{\mathrm{MCI}}$}
  & \textbf{AUC-ROC}
  & \textbf{Macro-F1}
  & \textbf{Accuracy}
  & \textbf{Recall$_{\mathrm{MCI}}$}
  & \textbf{Prec.$_{\mathrm{MCI}}$} \\
\midrule
Task-Specific Tokens (proposed)
  & $\mathbf{0.617} \pm 0.055$
  & $0.784 \pm 0.031$
  & $0.722 \pm 0.054$
  & $0.764 \pm 0.063$
  & $\mathbf{0.648} \pm 0.094$
  & $0.607 \pm 0.101$ \\[2pt]
Shared Learnable Token
  & $0.613 \pm 0.047$
  & $\mathbf{0.796} \pm 0.021$
  & $\mathbf{0.727} \pm 0.032$
  & $\underline{0.776} \pm 0.032$
  & $\underline{0.614} \pm 0.095$
  & $\underline{0.626} \pm 0.083$ \\[2pt]
CLS Token Only (no prompt)
  & $\underline{0.614} \pm 0.042$
  & $\underline{0.782} \pm 0.031$
  & $\underline{0.731} \pm 0.028$
  & $\mathbf{0.783} \pm 0.031$
  & $0.600 \pm 0.104$
  & $\mathbf{0.644} \pm 0.070$ \\
\bottomrule
\end{tabular}%
}
\end{table*}

Both the shared learnable token and the CLS-only configurations achieve
higher accuracy and, in the case of the shared token, higher AUC-ROC than
the task-specific design.  However, both alternatives yield lower
$F1_{\mathrm{MCI}}$ and MCI-class recall, indicating a systematic shift in
the operating point toward the healthy control class.  This behaviour is
consistent with a reduction in representational flexibility: a shared token
or a fixed CLS token must produce a single embedding that simultaneously
accounts for all three drawing modalities, which may cause the model to
anchor on the visually dominant healthy-pattern drawings, where all three
tests are executed correctly, at the expense of the subtler task-specific
deficits characteristic of MCI subjects.  The task-specific design, by
contrast, allows each learnable token to attend selectively to the
drawing-level features most relevant to its assigned test, preserving the
sensitivity required to detect partial or task-selective impairments. The task-specific configuration is therefore adopted as the proposed
baseline, as it achieves the best trade-off between MCI sensitivity and
overall discrimination, which is the criterion most consequential for a
clinical screening tool.

\subsection{Comparison of Parameter-Efficient Fine-Tuning Strategies}
\label{subsec:ablation_peft}

Having established task-specific prompt tokens as the preferred adaptation
signal, we compare prompt tuning against two alternative PEFT strategies:
top tuning and Low-Rank Adaptation
(LoRA)~\cite{hu2022lora}.  This ablation is conducted under the same
baseline conditions as Section~\ref{subsec:ablation_prompt}, with the
backbone frozen except where each strategy explicitly introduces trainable
parameters.

\paragraph{Top tuning} corresponds to the CLS-token-only configuration of
Section~\ref{subsec:ablation_prompt}: no learnable tokens are added and no
backbone weights are modified.  The pre-trained CLS token output is used
directly as the per-modality embedding, so the only trainable parameters
reside in the fusion module and the classifier.

\paragraph{LoRA} applies low-rank decomposition to the query, key, and value
projection matrices of every transformer block. Each frozen weight matrix $\mathbf{W}_0$
is adapted as $\mathbf{W} = \mathbf{W}_0 + \frac{\alpha}{r}\mathbf{B}\mathbf{A}$,
where $\mathbf{A} \in \mathbb{R}^{r \times d}$ and $\mathbf{B} \in \mathbb{R}^{d
\times r}$ are trainable low-rank factors with rank $r$ and scaling coefficient
$\alpha$. The CLS token output serves as the per-modality embedding, which is then
passed to the shared cross-attention and fusion modules. Two rank configurations are
evaluated: $(r{=}8, \alpha{=}16)$ and $(r{=}4, \alpha{=}8)$.


Table~\ref{tab:peft_ablation} reports $F1_{\mathrm{MCI}}$, AUC-ROC, and
MCI-class recall alongside the number of trainable parameters for each
strategy.  These three metrics are selected as sufficient to characterise the
primary performance objective, ranking calibration, and MCI sensitivity
respectively; together they capture the clinically relevant axes of
comparison without redundancy.

\begin{table}[t]
\centering
\caption{%
  Comparison of parameter-efficient fine-tuning strategies under baseline
  training conditions.  Results are mean $\pm$ std across five folds.
  \textbf{Bold} marks the best result per column;
  \underline{underline} marks the second best.%
}
\label{tab:peft_ablation}
\setlength{\tabcolsep}{7pt}
\renewcommand{\arraystretch}{1.30}
\begin{tabular}{l ccc r}
\toprule
\textbf{Configuration}
  & \textbf{$F1_{\mathrm{MCI}}$}
  & \textbf{AUC-ROC}
  & \textbf{Recall$_{\mathrm{MCI}}$}
  & \textbf{Trainable Params} \\
\midrule
Prompt Tuning (proposed)
  & $\mathbf{0.617} \pm 0.055$
  & $\underline{0.784} \pm 0.031$
  & $\mathbf{0.648} \pm 0.094$
  & \underline{1{,}188{,}098} \\[2pt]
Top Tuning
  & $\underline{0.614} \pm 0.042$
  & $0.782 \pm 0.031$
  & $\underline{0.600} \pm 0.104$
  & \textbf{1{,}186{,}946} \\[2pt]
LoRA ($r{=}4$, $\alpha{=}8$)
  & $0.610 \pm 0.068$
  & $\mathbf{0.787} \pm 0.042$
  & $0.624 \pm 0.140$
  & 1{,}260{,}674 \\[2pt]
LoRA ($r{=}8$, $\alpha{=}16$)
  & $0.605 \pm 0.026$
  & $0.778 \pm 0.027$
  & $0.550 \pm 0.073$
  & 1{,}334{,}402 \\
\bottomrule
\end{tabular}
\end{table}

Prompt tuning achieves the highest $F1_{\mathrm{MCI}}$ and MCI-class recall
of all four strategies while introducing only 1,152 parameters beyond those
of top tuning (three prompt tokens of dimension $d{=}384$).  This marginal
addition is sufficient to produce a measurable gain in MCI sensitivity,
demonstrating that task-dedicated input-space adaptation is more effective
per added parameter than modifying the backbone projections through low-rank
residuals.

The two LoRA configurations introduce between 72,576 and 147,456 additional
trainable parameters relative to prompt tuning, yet both yield lower
$F1_{\mathrm{MCI}}$ and recall.  The larger LoRA variant ($r{=}8$) achieves
the lowest recall (0.550) and highest standard deviation in accuracy across
all configurations, suggesting that introducing a larger number of trainable
parameters into the frozen backbone can destabilise training on a dataset of
this size, particularly under class imbalance.  The smaller variant ($r{=}4$)
is more stable but still underperforms prompt tuning on the primary metric.

These results collectively confirm that prompt tuning is the preferred PEFT
strategy for this task: it achieves the strongest MCI-class performance,
introduces the fewest additional parameters, and preserves the full
interpretability guarantees of the frozen backbone.
\section{Discussion}
\label{sec:discussion}

\paragraph{Parameter efficiency and representational transfer.}
A frozen DINOv2-Small backbone adapted through three prompt tokens of
dimension $d{=}384$ (1,152 backbone-stage parameters, under 6\% of the total
model) consistently outperforms ResViT~\cite{sirshar2026mci} across all
reported metrics, with a margin of 0.086 in MCI-class F1 and 0.043 in
AUC-ROC for the base model alone.  This confirms that self-supervised
patch-level representations transfer effectively to the neuropsychological
drawing domain without modifying backbone weights, and that architectural
complexity is not a prerequisite for strong performance on this task.  The
PEFT comparison (Table~\ref{tab:peft_ablation}) reinforces this point: LoRA,
which introduces up to 147,000 additional trainable parameters into the
backbone projections, yields lower MCI-class F1 and recall than prompt
tuning, suggesting that modifying the frozen attention structure on a dataset
of this scale is counterproductive.

\paragraph{Task-specific versus shared representations.}
The prompt-token ablation (Table~\ref{tab:prompt_ablation}) shows that
task-specific tokens preserve MCI sensitivity better than a shared token or
the frozen CLS token, even the latter two achieve marginally higher
accuracy and AUC.  This dissociation arises because aggregate metrics are
dominated by the majority class under a 2.4:1 imbalance, whereas MCI-class
F1 and recall quantify the sensitivity that matters clinically.
Shared or frozen queries must represent all three drawing types with a single
attention pattern, which likely anchors on the dominant healthy-subject
features common to all three modalities, at the cost of sensitivity to
partial, task-selective deficits.  Task-specific tokens are free to
specialize independently per drawing type, preserving diagnostic signal even
when only one modality carries evidence of impairment.

\paragraph{Interpretability as an architectural property.}
Both reference works treat interpretability as a post-hoc step:
Ruengchaijatuporn et al.~\cite{ruengchaijatuporn2022explainable} apply
attention rollout after inference, and Sirshar et al.~\cite{sirshar2026mci}
apply Grad-CAM~\cite{selvaraju2017grad} over the convolutional stream.  In
the proposed framework, the cross-attention weights from each prompt token to
the patch tokens constitute the saliency map; no additional computation is
required.  Because these weights are the same mechanism that produces the
task embedding, they directly reflect the information used during
classification rather than a gradient-based approximation thereof.  The
second interpretability layer, the modality importance weights
$\boldsymbol{\alpha}$, has no counterpart in either reference work and allows
inspection of which cognitive domain drove a given prediction, a capability
that is directly relevant to clinical decision support.

\paragraph{MoCA-adapted loss formulation.}
The leave-one-out ablation confirms that the MoCA-probability modulator is
the most consequential loss component ($-0.017$ in $F1_{\mathrm{MCI}}$ upon
removal), exceeding the contribution of MoCA soft labels ($-0.005$).  This
ordering holds in the individual strategy analysis as well, where the
modulator produces a larger isolated gain than the soft-label term.  The
soft-label approach of~\cite{ruengchaijatuporn2022explainable} encodes
score-derived uncertainty in the target; the proposed modulator encodes it
in the loss gradient, upweighting samples whose predicted probability
diverges from the score-derived expectation.  The two mechanisms address the
diagnostic boundary problem from complementary directions, and are most
effective in combination, as the full-system results demonstrate.  Adaptive
MoCA-bin weighting addresses a further dimension that neither reference work
considers: the non-uniform density of subjects across the MoCA range, which
concentrates under-represented, borderline subjects in sparsely populated
score bins.

\paragraph{Limitations and future directions.}
All experiments use a single dataset from one clinical institution in
Thailand with a specific demographic profile; generalizability to other
populations remains to be established.  The framework operates on static
drawing images and discards the temporal trajectory recorded by the digital
platform; kinematic features such as pen velocity, pressure, and stroke order
have been shown to carry independent diagnostic
value~\cite{davoudi2021classifying} and represent a
natural extension.  The modality importance weights $\boldsymbol{\alpha}$
have not been validated against clinical expert judgment; future work should
assess whether they align with the cognitive domains that practitioners
identify as most impaired in individual subjects.  Finally, replacing the
scalar MoCA score with domain-specific subscores could allow each
task-specific token to be guided by the cognitive dimension it is designed to
assess, potentially strengthening both performance and interpretability.









\section{Additional information}
\textbf{Code and data availability.} Code is available at \url{https://github.com/JVD9kh96/mci-detection}. Data is
publicly available at \url{https://github.com/cccnlab/MCI-multiple-drawings}.




\bibliographystyle{unsrt}  
\bibliography{references}

\end{document}